\documentclass[american,twoside, 11pt]{article}
\usepackage[T1]{fontenc}
\usepackage[latin9]{inputenc}
\usepackage{prettyref}
\usepackage{amsmath}
\usepackage{amssymb}
\usepackage{graphicx}
\usepackage[authoryear]{natbib}

\makeatletter
\usepackage[preprint,abbrvbib]{jmlr2e}
\usepackage{breakurl}
\jmlrheading{}{2022}{}{}{}{}{Christian Keup and Moritz Helias}
\ShortHeadings{Origami in N Dimensions}{Keup and Helias}
\firstpageno{1}

\newrefformat{fig}{\hyperref[#1]{Fig.~\ref{#1}}}
\newrefformat{sec}{\hyperref[#1]{Section~\ref{#1}}}
\newrefformat{app}{\hyperref[#1]{Appendix~\ref{#1}}}
\renewcommand{\textendash}{--}

\usepackage{microtype}

\makeatother

\usepackage{babel}
\begin{document}

\title{Origami in N dimensions: How feed-forward networks manufacture linear
separability}

\author{\name Christian Keup \email c.keup@fz-juelich.de \\
       \addr Institute for Computational and Systems Neuroscience (INM-6), \\
		Theoretical Neuroscience (IAS-6) and \\
		JARA Brain structure function relationships (INM-10), Research Centre Jülich, Jülich, Germany \\
		RWTH Aachen University, Aachen, Germany \\
       \AND
       \name Moritz Helias \email m.helias@fz-juelich.de \\
       \addr Institute for Computational and Systems Neuroscience (INM-6), \\
		Theoretical Neuroscience (IAS-6) and \\
		JARA Brain structure function relationships (INM-10), Research Centre Jülich, Jülich, Germany \\
		Department of Physics, RWTH Aachen University, Aachen, Germany}

\maketitle

\begin{abstract}
Neural networks can implement arbitrary functions. But, mechanistically,
what are the tools at their disposal to construct the target? For
classification tasks, the network must transform the data classes
into a linearly separable representation in the final hidden layer.
We show that a feed-forward architecture has one primary tool at hand
to achieve this separability: progressive folding of the data manifold
in unoccupied higher dimensions. The operation of folding provides
a useful intuition in low-dimensions that generalizes to high ones.
We argue that an alternative method based on shear, requiring very
deep architectures, plays only a small role in real-world networks.
The folding operation, however, is powerful as long as layers are
wider than the data dimensionality, allowing efficient solutions by
providing access to arbitrary regions in the distribution, such as
data points of one class forming islands within the other classes.
We argue that a link exists between the universal approximation property
in ReLU networks and the fold-and-cut theorem \citep{Demaine1998}
dealing with physical paper folding. Based on the mechanistic insight,
we predict that the progressive generation of separability is necessarily
accompanied by neurons showing mixed selectivity and bimodal tuning
curves. This is validated in a network trained on the poker hand task,
showing the emergence of bimodal tuning curves during training. We
hope that our intuitive picture of the data transformation in deep
networks can help to provide interpretability, and discuss possible
applications to the theory of convolutional networks, loss landscapes,
and generalization. 
\end{abstract}

\section{Introduction}

Trained neural networks are highly complicated functions that map
from the data space to a more useful representation space. Classical
proofs show that a feed-forward, multilayer architecture can approximate
any function in the limit of infinite layer widths \citep{Hornik1989,Cybenko1989,Funahashi1989,Barron1994}.
Yet given the architecture's expressivity, the question remains whether
the type of functions parameterized naturally by the architecture
allow a good solution to be found by the training procedure, and whether
it generalizes to unseen data samples. One way to study these questions
is via the loss landscape \citep[e.g. ][]{Gardner88_271,Gur-Ari18_04754,Abbara20_PMLR,DAscoli20_03509,Mannelli20_PRX,Lewkowycz20_02218}.
Here we start from a different perspective: We ask, what are the fundamental
operations available to the network to construct the output? 

We consider feed-forward, fully connected networks trained on classification
tasks. A focus on the ReLU nonlinearity yields a particularly intuitive
perspective, which can then be directly transferred to other activation
functions. Multilayer ReLU networks define piecewise-linear functions
where each border is related to the readout hyperplane of a neuron
\citep[see e.g. ][]{Raghu17_2847,Hanin19_00904,Balestriero2019_08443}.
The relative arrangement of these hyperplanes is sufficient to understand
the transformation implemented by the network (\prettyref{sec:Hammer-and-anvil}).

For a classification task, the last hidden layer of the trained network
must feature a representation of the data in which classes can be
separated by linear readouts from the output layer. Therefore, through
the layers, the network must progressively transform an initial, linearly
nonseparable data-distribution into a linearly separable form. What
types of transformation can a layer implement to increase the linear
separability of a representation, and which transformations are irrelevant?
We argue that there is one highly efficient mechanism to increase
linear separability, based on folds of the data manifold in unexplored
dimensions (\prettyref{sec:Folding-in-higher-dim}), while other mechanisms
that work also without additional dimensions are much less efficient.
In \prettyref{sec:Analyzing-the-transformation} we discuss how the
folding mechanism can be analyzed in high-dimensional real-world networks.

Our primary aim in this manuscript is to build a new intuition for
the processing inside neuronal networks. Therefore we abstain from
most mathematical analysis. However, our hope is that the intuition
presented here will be useful to construct new proofs dealing with
trainability and generalization.

\section{Forging the representation's shape: stereotypic nonlinearity and
affine transformation\label{sec:Hammer-and-anvil}}

\begin{figure}
\includegraphics{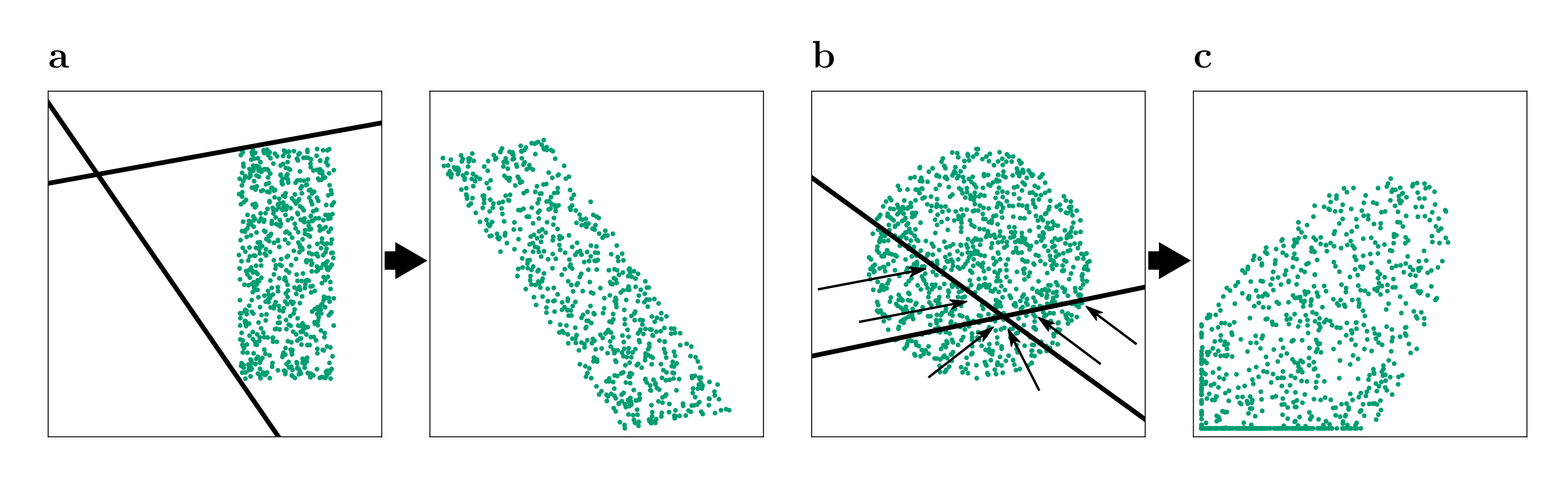}\caption{\textbf{Possible transformations of a data representation by a single
layer.} (a) Positioning on the anvil: Hyperplanes (black), whose
normal vectors and displacements are given by the rows of $W$ and
the biases $b$, define the affine transformation which has no effect
on separability, lines are mapped to lines. But it implements a rotation
and a positioning of the data distribution for the nonlinearity to
act on. (b) The hammer: Hyperplanes in a ReLU network can be thought
of as a large hammer making dents in the distribution. Only the convex
hull of the distribution is modified \textendash{} the architecture
cannot selectively target regions inside it. (c) The resulting representation
in the next layer. Figuratively, the distribution has been thrown
into the corner of an $N$-dimensional room.\label{fig:hammer-and-anvil}}
\end{figure}

The standard feed-forward architecture is a peculiar combination
of alternating affine transformations and an element-wise applied
nonlinear function
\begin{equation}
x_{i}^{(l)}=\Phi\left(W_{ij}^{(l)}x_{j}^{(l-1)}+b_{i}^{(l)}\right),
\end{equation}
with $x_{i}^{(l)}$ being the activation of neuron $i$ in layer $l$
and $\vec{x}^{(0)}$ the input data. The trainable weight matrix $W^{(l)}$
and bias vector $\vec{b}^{(l)}$define the affine transformation,
and $\Phi:\,\mathbb{R}\to\mathbb{R}$ is the nonlinear activation
function. We choose rectified linear units $\Phi(\tilde{x})=\mathrm{ReLU}(\tilde{x})=\max(0,\tilde{x})$,
but also discuss the effects of other activation functions. Preactivations
are denoted as $\vec{\tilde{x}}^{(l)}=W^{(l)}\vec{x}^{(l-1)}+\vec{b}^{(l)}$.
The data, being comprised of a number of classes, follows a ground-truth
probability distribution $\mathcal{P}_{x^{(0)}}$. What does the transformed
distribution $\mathcal{P}_{x^{(1)}}$ and those in successive layers
look like? How can the transformation promote linear separability
of the classes, which is required for classification in the output
layer?

The affine transformation itself cannot improve linear separability,
because all lines are mapped to lines. One way to picture the affine
transformation is to note that it can be uniquely specified by a mapping
between two arbitrary parallelograms. Another way is to picture an
arbitrary arrangement of hyperplanes, defined for each unit by $\tilde{x}_{i}\overset{!}{=}0$
(\prettyref{fig:hammer-and-anvil}a); then the normal vectors of these
hyperplanes constitute the new set of coordinate axes. Even though
a representation can be stretched, translated, sheared, rotated and
mirrored by an affine transformation, the resulting shape with respect
to linear readouts is always equivalent (\prettyref{fig:hammer-and-anvil}a).

Clearly, linear separability can only be improved by the nonlinearity,
but the parameters of the nonlinearity are typically fixed. Thus,
the network must use its freedom from the affine transformation to
position the representation such that the application of the nonlinearity
yields a beneficial deformation. The element-wise ReLU nonlinearity
can be visualized by $N$ hyperplanes separating positive preactivations
from negative preactivations; All data points on the negative side
of a hyperplane are projected onto the hyperplane since all negative
preactivations map to zero activation. Since each neuron corresponds
to one hyperplane, the combined effect is that of a corner of an $N$-dimensional
room into which the representation is pressed after application of
the affine transformation (\prettyref{fig:hammer-and-anvil}c). Therefore,
each ReLU can be thought of as a hammer-blow making a flat dent in
the distribution (\prettyref{fig:hammer-and-anvil}b), and the affine
transformation as the positioning (and scaling) of the distribution
on the anvil (\prettyref{fig:hammer-and-anvil}a). Other types of
nonlinearities have the same qualitative 'hammer' effect, only that
they do not necessarily map sets of points to a single hyperplane,
but result in a more gradual compactification of the points along
the direction of the normal vector. If the nonlinearity has a sigmoidal
shape, the compactification is applied from both sides.

As a consequence, the nonlinearity can only modify the convex hull
of the representation, because to affect a data point inside the distribution
all points further to the outside along one direction must be affected
as well. How can this transformation increase the separability of
classes within the distribution? In other words: To separate a set
of points belonging to one class that is at least partly surrounded
by points of other classes, how can concave dents in the representation
be realized? In the next section, we show how this is efficiently
possible as long as the layer provides higher dimensions which are
not explored by the data manifold. We also argue that this folding
mechanism is sufficient to provide the universal approximation property.
There are two further mechanisms we could identify which do not require
an expansion of dimensionality but are much less efficient, these
are discussed in \prettyref{app:Shear} and \prettyref{app:Folding-without-dimensionality-exp}.

\section{Folding in unoccupied, higher dimensions\label{sec:Folding-in-higher-dim}}

\begin{figure}
\includegraphics{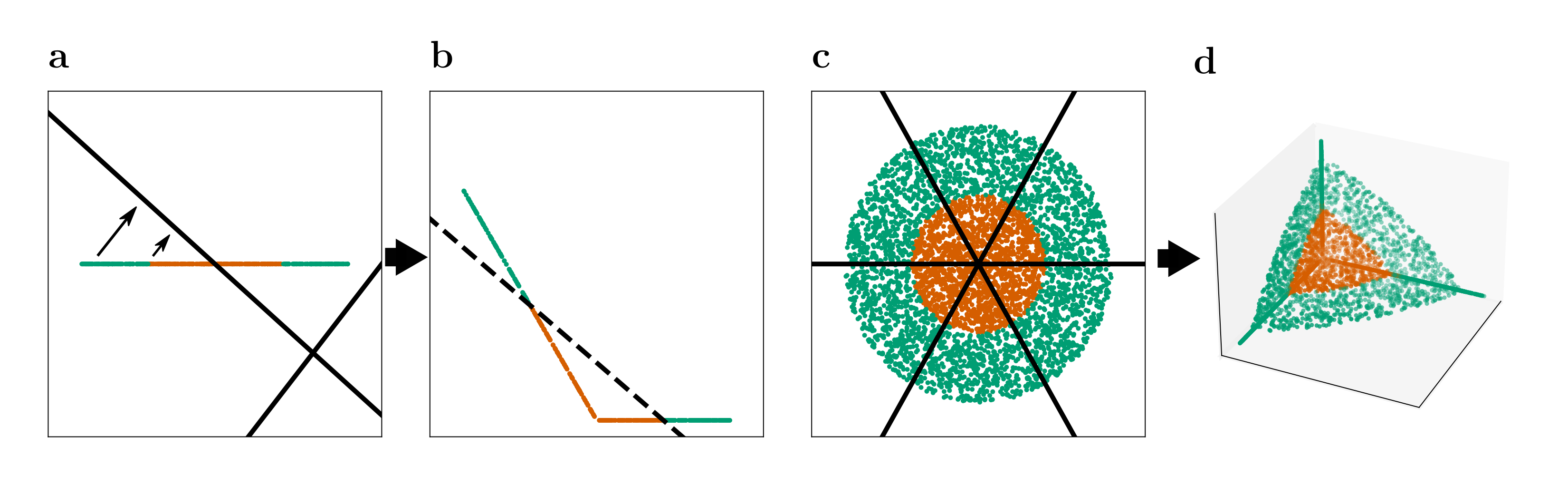}\caption{\textbf{Use of folding to efficiently expose an inner class boundary.}
(a) Fold of an ``1d-egg'' distribution, where the classes are the
inner and the outer part of the egg, induced by a ReLU hyperplane
with a component in the unexplored second dimension. (b) Separability
of the resulting representation by a linear readout. (c) Learned,
and also globally optimal hyperplane configuration solving the ``2d-egg''
problem by folding with 3 hidden neurons. (d) The corresponding folded
3d-representation in the hidden layer. Figuratively, the flat distribution
has been thrown into the corner of a 3 dimensional room. Recall of
a linear decision plane is $100\%$ for the inner and $96,8\%$ for
the outer class. A wider hidden layer would allow additional folds,
improving the roundish triangular approximation of the circle to a
roundish $N$-gon. \label{fig:Use-of-folding}}
\end{figure}

So far, in \prettyref{fig:hammer-and-anvil} we have considered a
situation in which the intrinsic dimensionality of the data manifold
equals the dimensionality of the layer. But the typical situation
is that the layers are wider than the input dimension, so in the first
layer at least, there are directions in activation space which are
not explored by the data distribution. This degeneracy has the advantage,
that if the hammer of a hyperplane hits the representation from an
unexplored direction, no datapoints are squashed together, they are
only displaced within this very coordinate. Such a mapping is nonlinear
but still injective. Therefore, the hammer blow of a ReLU hyperplane
that has a component into a direction in which the data manifold has
no extension results in a folding of the representation into that
direction (\prettyref{fig:Use-of-folding}a). Along the folding edge,
crucially, this exposes an internal region of the distribution to
the higher dimensional, embedding space. The exposed edge allows a
linear readout from the next layer to separate previously inaccessible
parts of the representation (\prettyref{fig:Use-of-folding}b). Note
that the angle between the normal vector and the representation subspace
must be larger than zero, but also smaller than orthogonal to create
a fold. To see how this principle generalizes, first consider the
stereotypic toy problem of linearly nonseparable classes, the ``2d-egg''
(\prettyref{fig:Use-of-folding}c), which can be (approximately) solved
using just one layer of three neurons (\prettyref{fig:Use-of-folding}c,d).
Concerning higher-dimensional distributions, the folding edge, which
becomes exposed, is always a hyperplane of one dimension less than
the dimensionality of the folded object (i.e. a line for a 2d object).
Compared to the mechanism based solely on shear discussed in \prettyref{app:Shear},
folding in unused dimensions is more efficient: Not only can several
folds be applied by a single layer, but the operation also scales
better to higher dimensions, as an $N$-dimensional egg problem can
be approximately solved by just one layer with $N+1$ neurons .

In a deep architecture each layer adds folds to the already folded
object, resulting in a hierarchical structure that can have exponentially
more edges than the total number of folds. A useful intuition for
this property is to think of folding an origami object: Most folds
are not applied in parallel, but progressively to the evolving object.
Folding is known to be a powerful operation: For 2-d sheets (paper)
it is proven that after appropriate folding, a single straight cut
is able to separate arbitrary shapes previously drawn on the sheet
(fold-and-cut theorem, \citealt{Demaine1998,Bern2011}). This result
is remarkably analogous to the case of neural networks, where arbitrary
classes can be separated from the (folded) data representation in
the last layer by a flat hyperplane. It is an enticing possibility
that the fold-and-cut theorem for 2d-sheets could be generalized to
the $N$-dimensional case and connected to the existing universal-approximation
theorems for neural networks.

We expect dimensionality expansion by folding operations throughout
a deep network to be the main resource to create separability, compared
to the additional mechanism provided by using shear (\prettyref{app:Shear}).
To prove this, a class of data distributions would need to be assumed.
However, an empirical test could be to restrict weights during training
to normal matrices, since the folding mechanism does not necessarily
require non-normal hyperplanes, and shear is thereby excluded. This
could be an advantageous architecture, as indicated by the results
on training very deep networks with orthogonal weight initialization
(leading to dynamical isometry, \citealt{Pennington17_04735,Xiao18_05393}).

Having identified folding as the basic operation a network can use
to achieve linear separability, we ask in the next section: How is
this mechanism used in a trained network, and how can folded structures
be analyzed in real-world networks that have very high dimensionality?

\section{Analyzing the transformation in trained networks\label{sec:Analyzing-the-transformation}}

The fundamental obstacle to understanding the transformation implemented
by real-world networks is high dimensionality, being a great challenge
to visualization, which is mostly confined to 3-d slices (linear techniques)
or 3-d manifolds (nonlinear techniques) of the N-d space. A simple
method would be to look at 2-d PCA plots of a layer, and plot lines
for a number of selected hyperplanes. To access folding, hyperplanes
can be selected that have both a component outside the span of the
representation as obtained by incremental PCA, and a significant proportion
of negative preactivations, which correspond to the data points affected
by the fold. This method may allow the exposure of feature generation
for simple data distributions, but is severely limited because it
assumes a connection between large variance and important features;
and because folds relying on several hyperplanes, for example the
solution of the 2d-egg problem in \prettyref{fig:Use-of-folding}c,
are not straight-forwardly detected by this method. However, it may
be usable in convolutional layers, because each filter is restricted
to a low-dimensional subspace, such as $3x3\,\mathrm{conv}\,=9\mathrm{d}$.

An approach that can overcome the limits of visualization is to define
observables which are indicative of specific operations, and measure
their occurrence. The dimensionality of the representation is an observable
related to folding. By doing a PCA on both the preactivations and
the activations of a layer, it can be accessed how many and which
directions have acquired nonzero variance, that is, have been folded
(\prettyref{fig:Analysis-of-poker-task}b). Because folds expose the
data at the kink, their use to create separability must mean that
these neurons respond little to the inner class, and at the same time
not at all or strongly to the outer classes (compare \prettyref{fig:Use-of-folding}).
A fold that improves linear separability should therefore go hand-in-hand
with a stereotypic bimodal tuning curve for the outer classes. Since
ReLU units map all negative inputs to zero, they cause a delta peak
in the tuning curves at $0$. So it is more informative to look at
the preactivations' tuning curves instead, which should show a dip
for a subset of classes close to zero. For neurons with a smoother
nonlinearity, also the activation tuning curves can be considered
instead. Analyzing this type of tuning as an observable thus indicates
the prevalence of linear separability-generation by folding also in
real-world trained networks (\prettyref{fig:Analysis-of-poker-task}c,d,
\prettyref{fig:Examples-of-tuning-curves}). \citet{Recanatesi2019_00443}
have shown that the dimensionality of the data representation in deep
networks often decreases in the last layers, while it increases throughout
the previous layers. We hypothesize that the bimodal type of tuning
should build up towards the ``completed fold'' in the layer of highest
representation dimensionality, and if the dimensionality is reduced
in the following layers, such bimodal tuning of neurons should be
reduced, because it is only beneficial in conjunction with dimensionality
expansion, but becomes detrimental to the conservation of separability
when dimensionality is compressed (see supplementary figures S3,
S4, S5 for a first validation).

Lastly, we propose to consider the angle between the normal vector
defining the hyperplane of a neuron and the subspace of the data representation.
Units that create folds (\prettyref{sec:Folding-in-higher-dim}) should
have an intermediary angle between $0$ and $\pi/2$. If they contribute
to the generation of separability, these units should furthermore
show the bimodal type of tuning curves. Randomly initialized units
should have an angle determined by the ratio of representation dimensionality
and layer width. For very wide layers, random angles would thus cluster
close to $\pi/2$, which might allow to distinguish neurons with random
angle from neurons whose angle was optimized during training to contribute
to linear separability generation.

\subsection{Poker hand task\label{subsec:Poker-hand-task}}

\begin{figure}
\includegraphics{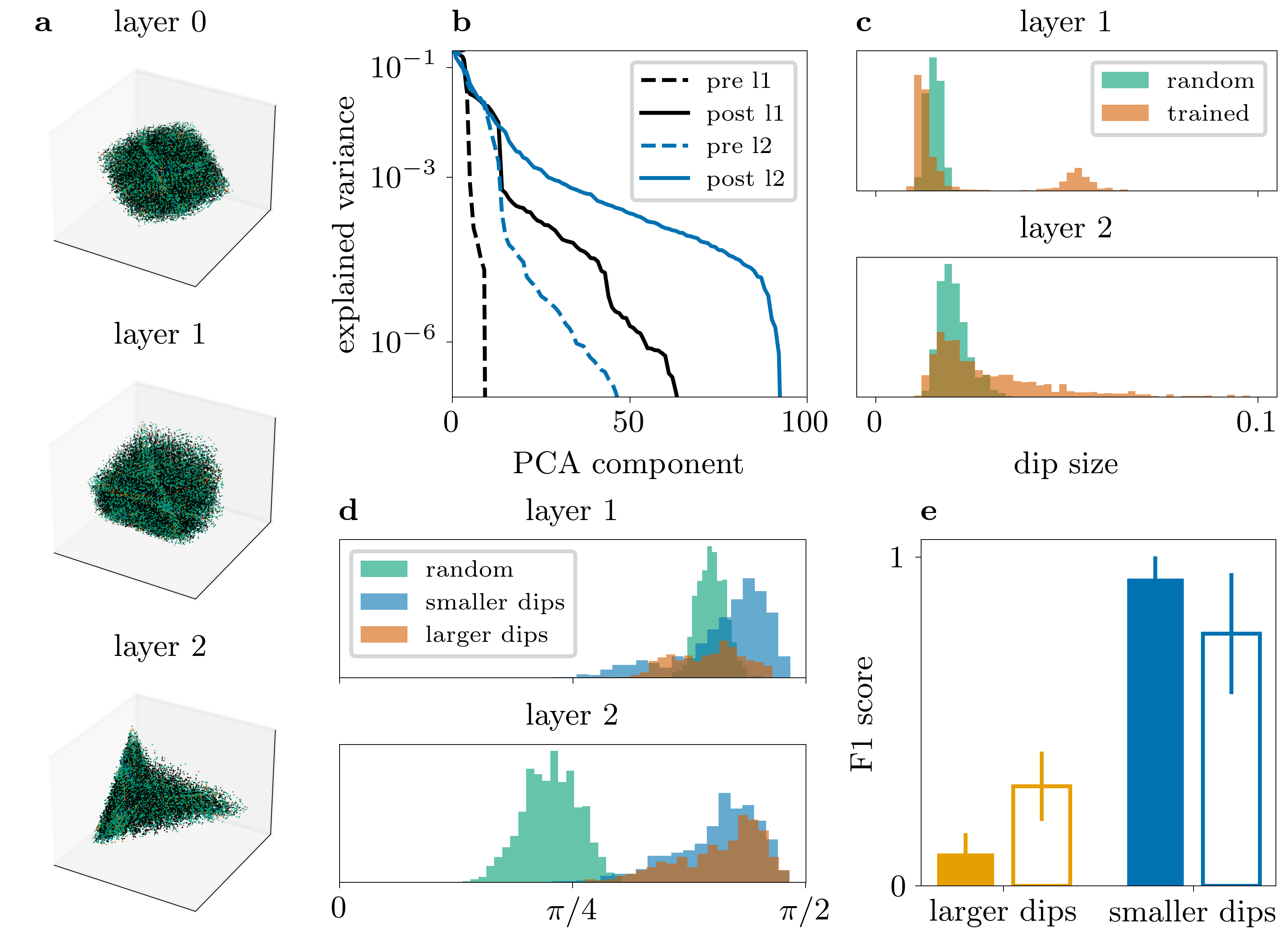}\caption{\textbf{Analysis of the transformation learned to solve the poker
hand task.} (a) 3d-PCA plots of the representation. Even though the
data is only 10 dimensional, the 3d-PCA is not very informative.
(b) Evolution of representation dimensionality in the two hidden layers,
showing strong folding related dimensionality expansion. (c) Deviation
from unimodality of preactivation tuning curves in layer 1 and 2 for
class 2, assessed by histogram of the Hartigan dip statistic for
each neuron. Surrogates (green) use a random layer 1 or 2, respectively.
Here and in panels (d,e) results are aggregated across 10 trained
network realizations. (d) Histogram of angles between the normal
vectors and the representation subspace, for neurons with large (dark
orange) and small dip statistic (blue). Random vectors (green) for
comparison. A neuron is classified as having a large dip if the dip
statistic for at least one class is larger than the maximal dip statistic
for the random layer surrogate.  (e) Large accuracy loss after silencing
$10$ neurons from the subset with larger dips, compared to $10$
neurons with smaller dip statistic. Silenced neurons from layer 1
(filled bars) and layer 2 (empty bars). Accuracy measured as macro-averaged
F1 score, excluding classes $\{5,8,9\}$. Mean score of the intact
networks is $0.97$.\label{fig:Analysis-of-poker-task}}
\end{figure}

We here analyze networks trained on the poker hand task from the UCI
repository \citep{Dua17_UCI_ML_repo}. The input data is 10 dimensional,
describing a five card poker hand (suit and rank each), and the targets
are $10$ classes constituting the type of hand (nothing, pair, two
pairs, three of a kind, straight, flush, full house, four of a kind,
straight flush, royal flush). Note that the later classes are much
less frequent and harder to learn, so we focus here on the first 7
classes. This task is well suited for our purpose for several reasons:
First, a linear classifier on this data has very low performance close
to chance, so the network must use the mechanisms of separability
generation to improve performance. Second, the data does not offer
an advantage to convolutional architectures, allowing us to restrict
the analysis here to the case of fully connected networks. Third,
since we are not interested in the generalization properties of the
trained networks, we can use the exhaustive data set for training,
which allows better interpretability because the true task structure
is learned. Lastly, the task is high-dimensional and complex enough
to defy visualization of the data structure (\prettyref{fig:Analysis-of-poker-task}a),
while still allowing a theoretical understanding.

\begin{figure}
\includegraphics{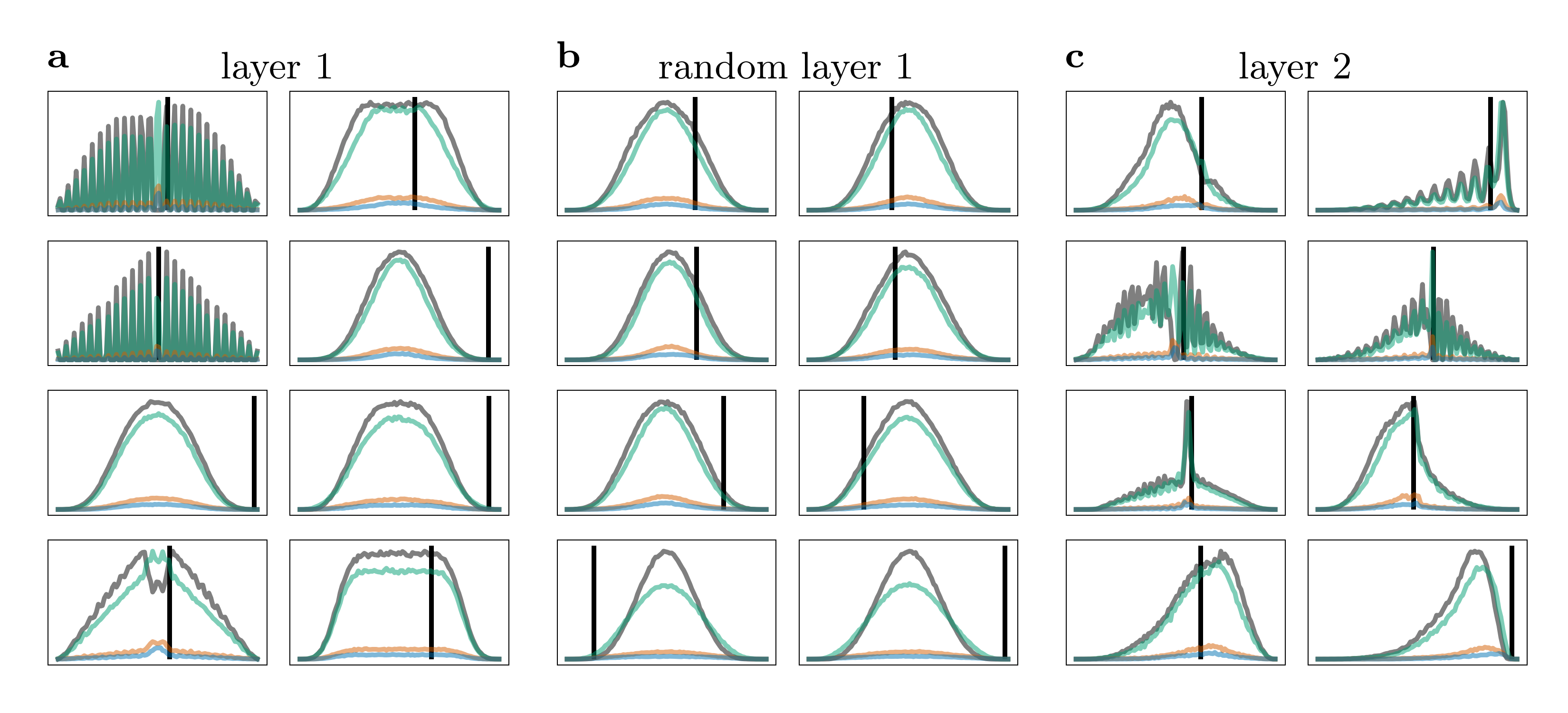}\caption{\textbf{Examples of preactivation tuning curves.} (a) Class-resolved
tuning curves (histograms) for the preactivations of a representative
subset of neurons in layer 1 after training. Vertical black line marks
the position of the folding edge at $\tilde{x_{i}}=0$. Colors of
the classes are 0/nothing (black), 1/pair (green), 2/two pairs (dark
orange), 3/triple (blue). (b) Close to Gaussian preactivation tuning
curves in untrained network. (c) Analogous to (a), but for the trained
layer 2. See figures S1 and S2 for the complete set of tuning curves.\label{fig:Examples-of-tuning-curves}}
\end{figure}

We train fully connected 3-layer ReLU networks with layer widths $\{10\rightarrow100\rightarrow100\rightarrow10\}$
using cross-entropy loss with momentum $0.9$, batchsize $500$ and
a learning rate schedule of $0.1$ for $50$ epochs then $0.01$ for
$50$ further epochs, implemented in pytorch. Although not in the
overparametrized regime, having $\sim1.2\times10^{4}$ parameters
compared to $\sim10^{6}$ training samples, the network shown in \prettyref{fig:Analysis-of-poker-task}
and \prettyref{fig:Examples-of-tuning-curves} achieves $100\%$ training
accuracy on all but the difficult classes $\{5,8,9\}$ (see suppl.
tables S6). Visualization of the data representation by means of the
first three principal components is of little use (\prettyref{fig:Analysis-of-poker-task}a).

An incremental PCA of the pre- and postactivations in each layer shows
a strong increase of dimensionality due to folding (\prettyref{fig:Analysis-of-poker-task}b).
Also the class specific tuning curves of the trained network show
the predicted sign of separability generating folds: Many of the tuning
curves are bimodal (or multi-modal), with a change of the dominating
class close to the folding edge (\prettyref{fig:Examples-of-tuning-curves}a,c),
see suppl. figures S1 and S2 for the curves of all 200 neurons),
this is in clear contrast to the corresponding tuning curves in a
random network before training, which are close to Gaussian (\prettyref{fig:Examples-of-tuning-curves}b).
The oscillating multi-modal nature of some of the trained tuning curves
is due to the categorical (discrete) nature of the data. The important
feature, however, is that in these curves the central peak close to
the folding edge is missing for some classes while existing for others,
in line with the prediction of bimodality (see also figures S1,
S2). Using Hartigan's dip statistic \citep{Hartigan85,Freeman12_83},
an elegant non-parametric measure of deviations from unimodality,
we find that the distribution of dip sizes across neurons clearly
tends to higher values in the trained compared to the untrained networks
(\prettyref{fig:Analysis-of-poker-task}c).

Neurons with large dip size for at least one class define hyperplanes
whose angle between normal vector and representation subspace tends
towards intermediary values (\prettyref{fig:Analysis-of-poker-task}d),
as expected from the theory; However, the angle distribution does
not show a difference to neurons with small dip sizes. A reason for
this could be, that the layers are not very wide compared to the representation
dimensionality so that also random angles fall into the intermediary
range, and these may not be penalized by the loss function because
unnecessary folds in unused dimensions are not detrimental to linear
separability. It is striking that in the second layer, the angles
in the trained networks have all shifted to higher values compared
to the untrained state (\prettyref{fig:Analysis-of-poker-task}d).
Note that no angle between normal vector and representation subspace
is close to zero, therefore, the mechanism of separability generation
solely by shear (\prettyref{app:Shear}) does not play a role in this
network.

Finally, while the angles to the representation subspace do not distinguish
between neurons with large and small dip statistics, those with large
dips are more important for the network function: Silencing $10$
neurons with large dip size results in a considerable loss of accuracy,
while silencing $10$ neurons with small dip sizes has a small effect
(\prettyref{fig:Analysis-of-poker-task}e).

\section{Conclusion\label{sec:Conclusion}}

In this work, we have developed an intuition for the type of transformations
a feed-forward network applies to the data distribution, drawing an
analogy to forging and folding. We found that in the context of classification,
the network has one basic tool at hand to efficiently manufacture
the linear separability of classes necessary in the final layer: To
progressively fold the distribution in higher, unexplored dimensions,
thereby efficiently exposing arbitrary internal regions of the distribution
to the outside. For ReLU networks this can be likened to folding an
$N$-dimensional origami object. For smooth nonlinearities, the folding
corresponds to a smooth bending. This does not make a qualitative
difference for classification: The bent region is exposed to the outside,
just like the region around a fold.

Although visualization in high-dimensional spaces is strongly limited,
the action of the folding mechanism can be analyzed also in large
realistic networks by defining observables causally linked to the
operations. Specifically, by using the insight into generation of
linear separability by folding, we found that mixed selectivity with
bimodal tuning curves, in which neurons are activated weakly by some
class, but at the same time not at all and strongly by another class
(or classes), are a causal sign of beneficial folding operations.
For very wide networks, also the angle between the weight vector of
a neuron and the subspace explored by the data representation might
be informative.

\subsubsection*{Limitations}

The work represents first steps towards an alternative understanding
of the highly complex transformations implemented by trained networks.
It is so far restricted to fully connected, deep feed-forward networks
and classification tasks. The statement about the relative inefficiency
of separability generation without dimensionality expansion is not
quantified by a rigorous calculation. That is because a proof would
require assumptions about the class of realistic data distributions,
an understanding of which is currently, to our knowledge, lacking.
However, that there are hard restrictions on the expressivity of networks
without dimensionality expansion is consistent with \citep{Johnson18_00393},
where it is proven that deep but fixed width networks are not universal
approximators. Finally, the poker hand task, on which we have presented
the initial validation of our predictions, is relatively small scale;
an extension of the theory to convolutional architectures would be
desirable to allow validation on large image classification models.
Also, the task data are categorical variables, which makes the analysis
of bimodality in the trained tuning curves less obvious due to the
intrinsic multi-modal structure along the original data axes.

\subsubsection*{Related work}

That ReLU networks are piecewise-linear functions is the basis of
a substantial body of literature \citep[e.g. ][]{Arora2016_01491,He2018_03973,Hanin19_00904,Cosentino2020_09525,Lakshminarayanan2021_03403,Zavatone21_arxiv}.
A notable approach \citep{Balestriero2018_06576,Balestriero2019_08443}
provides insight into ReLU architectures by linking them to affine
spline operators and showing that the networks perform a greedy template
matching. Concerning works that study the representation manifold,
mathematical frameworks constructing the mapping of smooth manifolds
through deep networks have been proposed in \citep{Benfenati2021_09656,Benfenati2021_10583}
and \citep{Hauser2017_NIPS}. The work \citep{Zhang2018_5824} shows
that for ReLU units, the manifold can be studied from the viewpoint
of tropical geometry. Closely related to our approach are \citep{Zhang2020_ICLR},
which investigates the properties of linear regions and hyperplane
arrangements, and \citep{Rolnick2019_00744,Carlsson2017_ICLR,Carlsson2019_08922,Gamba2019_ICCVW}
studying the preimage of classification boundaries arising in ReLU
feed-forward networks. These works are not explicitly concerned with
the link between folding and linear separability, however. In \prettyref{sec:Folding-in-higher-dim}
we have argued that the analogy between ReLU networks and (high-dimensional)
paper folding goes so far that the fold-and-cut theorem \citep{Demaine1998,Bern2011}
corresponds to a universal approximation theorem based on paper folding.
At least for convex polyhedra, this theorem has been generalized to
arbitrary dimensions by a proof that convex polyhedra are flat-foldable
\citep{Abel2014_396}. Wide ReLU networks which have arbitrary many
ambient dimensions can relax the requirement that the linear elements
of the decision boundary are to be folded exactly on top of each other.
Therefore we hypothesize that these approaches can be used to show
that the folding operations in ReLU networks can create arbitrary
(non-convex) piecewise-linear decision boundaries, but here leave
this to future work. 

\subsubsection*{Benefits of increasing the mechanistic intuition for deep networks}

We hope that our perspective on the generation of linear separability
by folding operations can inspire new theoretical approaches to the
questions of trainablity and generalization of network architectures.
For example, it could be investigated how the training gradients move
the hyperplanes, in order to gain an understanding of the loss landscape
and its local minima. The mechanistic understanding of separability
generation could also lead to more powerful architecture parametrizations
that allow one to naturally learn meaningful operations on the data
representation. For overparameterized models it would not help to
render the loss landscape more convex, because it is not sufficient
to find a minimum that possibly overfits the data, but instead a minimum
is needed that in addition guarantees good generalization. The generalization
properties are given by the prior of an architecture, which in turn
is tightly related to the type of operations it tends to learn.

From a neuroscientific perspective, we have presented a principled
argument linking mixed selectivity and bimodality of tuning curves
to the solution of classification tasks. This is relevant because
traditionally neuroscientists have tended to search for mono-modal
tuning curves strongly selective for a single class \citep{Kandel13,DiCarlo2007_333};
At the same time, experimentally observed tuning curves are notoriously
complex, with mixed selectivity being the rule \citep{Rigotti2013_585,Parthasarathy2017_1770}.
It is necessary to distinguish between a layer with a linearly separable
representation, in which mixed selectivity is to be expected, and
a readout from such a layer, which can be strongly selective.

Good directions for further work would be extensions to convolutional
networks, transformers, and to incorporate the common practice of
dropout. For convolutional layers, it should be possible to think
of each filter patch as a small fully connected network. It then
follows that one needs more filters than the number of dimensions
explored by the data under the filter patch, otherwise no folding
is possible and the extractable features are largely linear. Another
avenue for future work would be to study the high-dimensional limit
and thereby add to the understanding of results from approaches to
feed-forward networks based on statistical mechanics \citep{Chung18_031003,Lee18,Cohen21_023034,Fischer2022_arxiv_04925,Bahri20_501}.

\subsubsection*{Acknowledgments}

This work was supported by the German Federal Ministry for Education
and Research (BMBF Grant 01IS19077A to Jülich), the Exploratory Research
Space (ERS) seed fund neuroIC002 (part of the DFG excellence initiative)
of the RWTH university and the JARA Center for Doctoral studies within
the graduate School for Simulation and Data Science (SSD), funded
by the Deutsche Forschungsgemeinschaft (DFG, German Research Foundation)
- 368482240/GRK2416, and funded by the Excellence Initiative of the
German federal and state governments (G:(DE-82)EXS-PF-JARA-SDS005).

\newpage{}

\appendix

\part*{Appendix}

\section{Pushing aside pieces of the distribution by shear\label{app:Shear}}

\begin{figure}
\includegraphics{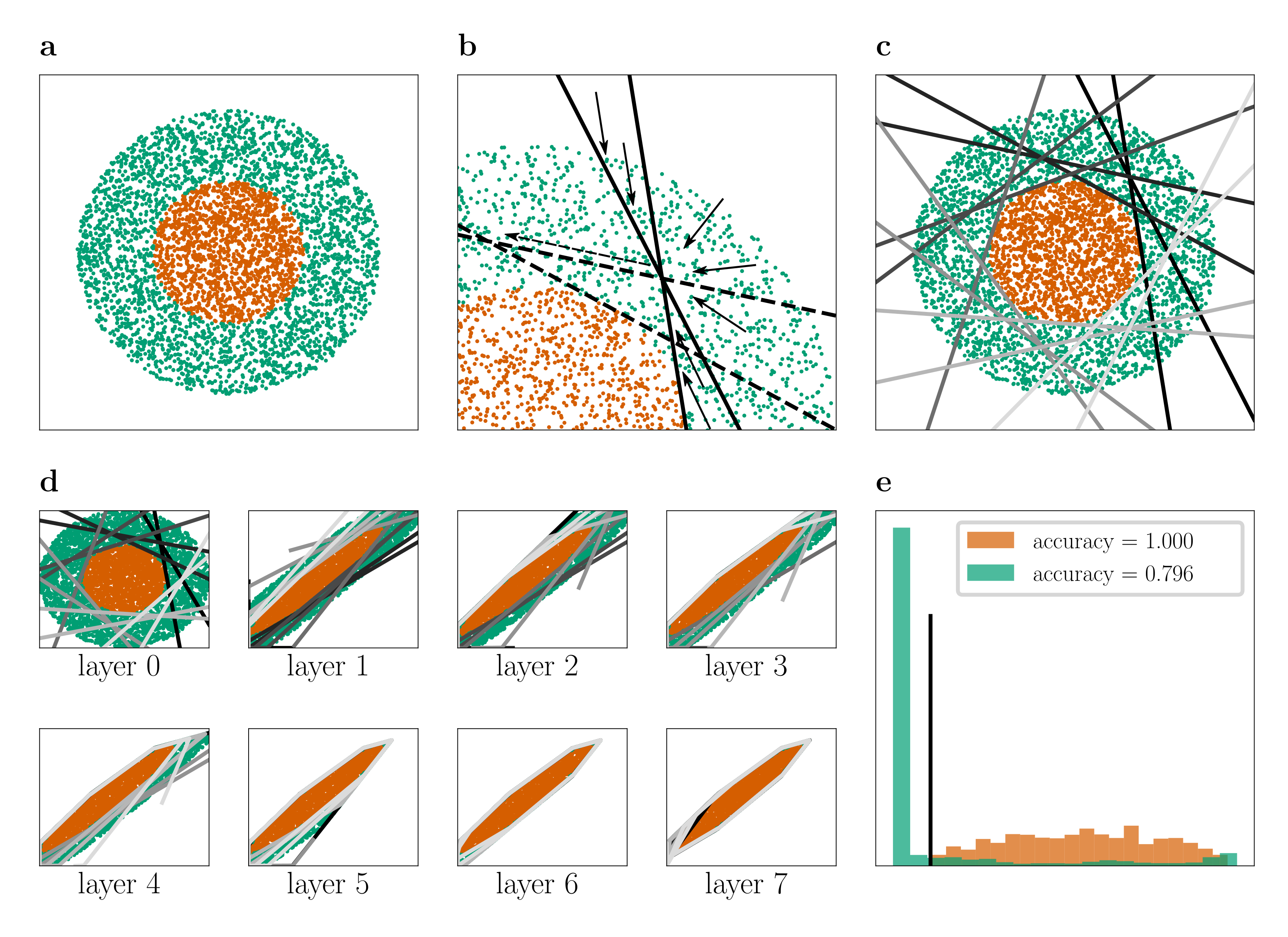}\caption{\textbf{Use of shear to expose an inner class boundary.} (a) The classic
``2d-egg'' toy problem of linearly nonseparable classes. (b) Method
of moving probability mass to the side in a deep network restricted
to width $2$, by interplay of nonlinearity and non-orthogonal hyperplanes
(shear). The majority of the squashed probability mass is concentrated
at the new origin, which is transported to the side by the next layer
(dashed lines). (c) 1d-hyperplane configuration of a $7$ layer ReLU
network of width $2$ that solves the 2d-egg problem, shown in the
input space. There are two lines per layer, from the first (black)
to $7$-th layer (white). (d) Corresponding evolution of the representation
through the layers. (e) Histogram of linear readout from last layer
showing linear separability of the two classes. Accuracies measured
as recalls in relation to the black decision line. \label{fig:Use-of-shear}}
\end{figure}

Consider again the ``2d-egg'' toy problem (\prettyref{fig:Use-of-shear}a).
As shown \prettyref{fig:hammer-and-anvil}, it is not trivial to access
the inner class because the hammer of a hyperplane cutting through
the distribution would only push the outer points inside as well.
Without using wider layers supplying additional dimensions (\prettyref{sec:Folding-in-higher-dim}),
we found only one method to solve the task: By using the fact that
the ReLU operation moves points in different directions, depending
on how many hyperplanes are crossed, it is possible to use non-normal
hyperplanes to iteratively cut pieces from the distribution, pushing
the probability mass to the side, thereby exposing the class boundary
inside the distribution, see \prettyref{fig:Use-of-shear}b. In this
way, the shell given by the outer class can be shaved off (\prettyref{fig:Use-of-shear}c,d),
resulting in a linearly separable representation in the last layer
(\prettyref{fig:Use-of-shear}e). It must be noted that only one such
operation is possible per layer; therefore many layers are required
to solve even the simple 2-dimensional toy problem. In higher dimensions,
the number of layers needed by this mechanism scales even worse, as
seen by considering a ``3d-egg'' problem (that is, a boiled egg
instead of a fried egg): Each layer can only push aside one slice
of the outer class, and the accumulated probability mass can only
be transported along a 1d-path. Thus it is not sufficient to go around
the sphere once as in (\prettyref{fig:Use-of-shear}c), but to trace
a spiral around it (as one would peel an orange). In fact, the restriction
that the cut needs to start from one point and proceed along one direction
is related to the result that deep, but fixed width networks are not
universal approximators \citep{Johnson18_00393}. Lastly, it is interesting
to note that in the solution presented in (\prettyref{fig:Use-of-shear}c,d)
each layer performs the exact same transformation, meaning that the
weights and biases are identical across layers. The network therefore
corresponds to a recurrent network which has been unrolled in time,
as done for BPTT \citep{Pearlmutter89}. Indeed, the equivalent 2-neuron
recurrent network also solves the task, see suppl. figure S7.

\section{Compressive folding without dimensionality expansion\label{app:Folding-without-dimensionality-exp}}

We can also consider the case where the hyperplanes are placed in
the same way as in \prettyref{sec:Folding-in-higher-dim}, but without
unexplored ambient dimensions. If the layer would provide unexplored
dimensions, this could also be seen as the case where the hyperplane
is orthogonal to the subspace occupied by the data manifold. Then
the data points on the negative side of the hyperplane are squashed
together as in \prettyref{fig:hammer-and-anvil}b/c and the representation
looses information. However this flat dent on the distribution can
straighten out one angle in a (piecewise linear) class boundary: If
the hyperplane goes through the intersection of the two linear elements
forming the angle, the effect is that one of the elements is reduced
to length zero (along the direction defined by the angle), therefore
the angle is gone and only a linear boundary remains. This operation
can be applied $N$ times per layer, as the folding operations in
\prettyref{sec:Folding-in-higher-dim}. Also, considering one hyperplane,
it is possible to chain the operation by using multiple layers, which
corresponds to applying squashing along a piecewise linear path. However,
because the mapping is not injective (in contrast to the case including
dimensionality expansion), there are limits on what kind of boundaries
can be created: (1) Any single angle along the piecewise linear path
of squashing must be smaller than $\pi/2$, and (b) the sum of angles
in any direction along the path is strictly bounded from above by
$\pi$. For example, in 2-d it is possible to approximate a half-circle
as the class boundary by using successive tangential squashes, but
to approximate a complete circle as done in \prettyref{fig:Use-of-folding}b,c
and also \prettyref{fig:Use-of-shear} is not possible.

\end{document}